\documentclass[10pt,twocolumn,letterpaper]{article}

\usepackage{iccv}
\usepackage{times}
\usepackage{epsfig}
\usepackage{graphicx}
\usepackage{amsmath}
\usepackage{amssymb}

\usepackage{booktabs}

\usepackage{multirow}
\usepackage{color}
\usepackage{rotating}
\usepackage{float}
\usepackage{stfloats}
\usepackage{subfigure}
\usepackage{ragged2e}
\usepackage{bm}

\usepackage{bbding}

\usepackage{pifont}

\usepackage[pagebackref=true,breaklinks=true,letterpaper=true,colorlinks,bookmarks=false]{hyperref}

\iccvfinalcopy 

\def\iccvPaperID{9372} 

\ificcvfinal\fi

\begin{document}


\title{Improving Zero-Shot Generalization for CLIP with Synthesized Prompts}

\author{Zhengbo Wang$^{1,2}$, Jian Liang\thanks{Corresponding author} \ $^{2,3}$, Ran He$^{2,3}$, Nan Xu$^{5}$, Zilei Wang$^1$, and Tieniu Tan$^{2,3,4}$ \\
$^1$ University of Science and Technology of China\\
$^2$ CRIPAC \& MAIS, Institute of Automation, Chinese Academy of Sciences\\
$^3$ University of Chinese Academy of Sciences
$^4$ Nanjing University $^5$ Beijing Wenge Group \\
{\tt\small zhengbowang@mail.ustc.edu.cn, liangjian92@gmail.com}
}

\maketitle
\ificcvfinal\thispagestyle{empty}\fi

\begin{abstract}

    With the growing interest in pretrained vision-language models like CLIP, recent research has focused on adapting these models to downstream tasks.
    Despite achieving promising results, most existing methods require labeled data for all classes, which may not hold in real-world applications due to the long tail and Zipf's law.
    For example, some classes may lack labeled data entirely, such as emerging concepts.
    To address this problem, we propose a plug-and-play generative approach called \textbf{S}ynt\textbf{H}es\textbf{I}zed \textbf{P}rompts~(\textbf{SHIP}) to improve existing fine-tuning methods.
    Specifically, we follow variational autoencoders to introduce a generator that reconstructs the visual features by inputting the synthesized prompts and the corresponding class names to the textual encoder of CLIP.
    In this manner, we easily obtain the synthesized features for the remaining label-only classes. 
    Thereafter, we fine-tune CLIP with off-the-shelf methods by combining labeled and synthesized features.
    Extensive experiments on base-to-new generalization, cross-dataset transfer learning, and generalized zero-shot learning demonstrate the superiority of our approach. 
    The code is available at \url{https://github.com/mrflogs/SHIP}.
    
\end{abstract}

\section{Introduction}

In recent years, language-supervised vision pretrained models have garnered much attention. 
By establishing a link between images and natural language, these models exhibit impressive zero-shot capabilities and remarkable transfer ability~\cite{radford2021learning, jia2021scaling, alayrac2022flamingo, li2022blip}, demonstrating potential in learning open-world concepts.
One of the most successful large-scale pretrained vision-language models is CLIP~\cite{radford2021learning}. 
By leveraging a massive dataset of 400 million image-text pairs, it learns to align visual and textual representations from a vision encoder and a language encoder, respectively.
After pretraining, CLIP~\cite{radford2021learning} can perform zero-shot recognition by merely providing the class names. 
The classification weights are generated by the language encoder through prompting~\cite{liu2023pre}. 
For instance, we can adopt a prompt template like ``a photo of a \{class\}" as the input of the text encoder, and then the weights for classification can be synthesized by substituting in the ``\{class\}" with the actual class name. 
And the resulting classification score is the cosine similarity between the test image and the weights. 

To further enhance the performance of CLIP, several previous works have proposed the use of learnable prompts~\cite{zhou2022learning, zhou2022conditional, derakhshani2022variational, lu2022prompt} or adapters~\cite{zhang2022tip, gao2021clip} to fine-tune the pretrained CLIP to specific downstream tasks. 
These methods have achieved significant improvements with only a small amount of labeled data from downstream datasets, which clearly demonstrates their superiority in terms of data efficiency.
However, a significant limitation of these methods is their reliance on having data available for all classes, which can be impractical in real-world applications.  
The issue arises due to Zipf's law and the long tail phenomenon, which make it challenging to collect data for rare categories, such as new species or emerging concepts. 
As a result, many categories may be devoid of any relevant data, rendering previous methods either invalid~\cite{wang2022learning, zhang2022tip} for such scenarios or leading to a significant drop in performance on the label-only classes~\cite{zhou2022learning}, compared to zero-shot CLIP. 
To address this limitation, our goal is to develop a fine-tuning approach that can effectively recognize both categories with and without available data while maintaining the superior data efficiency of previous methods. 

In this paper, we propose a plug-and-play generative approach called \textbf{S}ynt\textbf{H}es\textbf{I}zed \textbf{P}rompts~(\textbf{SHIP}) to improve existing fine-tuning methods. 
The main objective is to train a generative model that can synthesize features by providing class names, which enables us to generate features for categories without data.
And we proceed to fine-tune CLIP using both the original labeled and the newly synthesized features with off-the-shelf methods.
However, a major obstacle is that generative models typically require a substantial amount of data to train, which contradicts our goal of data efficiency.
We propose to utilize variational autoencoder~\cite{kingma2013auto} (VAE) as the framework, which is easier to train and more effective in low-data scenarios compared to models that require adversarial training~\cite{arjovsky2017wasserstein, goodfellow2020generative}.
Additionally, inspired by previous prompt learning methods~\cite{zhou2022learning, zhou2022conditional, derakhshani2022variational, lu2022prompt}, we train the generator to produce prompts instead of visual features.
We then feed these prompts and corresponding class names into the frozen CLIP language encoder to obtain synthesized features.
Since CLIP has been pretrained on a large-scale dataset and has aligned visual and language representations, we believe that the pretrained language encoder aids in generating more realistic features. 

In summary, this paper aims to address the issue of downstream tasks where some classes have no relevant data while maintaining the superior data efficiency of previous methods.
To achieve this goal, we propose a novel generative approach named SHIP, which can synthesize features for categories without data based solely on their class names.
Notably, our proposed generative method is orthogonal to CLIP fine-tuning methods and can enhance their performance by utilizing synthesized data.
We conduct comprehensive experiments on base-to-new generalization, cross-dataset transfer learning, and generalized zero-shot learning, resulting in state-of-the-art performance.

\section{Related Work}

\noindent\textbf{Vision-Language Pretraining.}
Vision-language pretraining models (VLMs) investigate the relationship between vision and language modalities.
Various methods have been proposed to establish this connection through self-supervised learning, such as masked language model~\cite{kim2021vilt, lu2019vilbert}, masked region prediction~\cite{tan2019lxmert, su2019vl} and image-text matching~\cite{tan2019lxmert, kim2021vilt}.
Recently, contrastive learning-based VLMs have shown remarkable performance by utilizing large-scale noisy image-text pairs.  
These methods, including CLIP~\cite{radford2021learning} and ALIGN~\cite{jia2021scaling}, learn aligned representations of images and text via the contrastive loss, which pulls the representations of matching image-text pairs together and pushes those of mismatching pairs apart.
Based on natural language supervision, these VLMs acquire transferable visual representations and exhibit impressive zero-shot performance on various image classification tasks.

\noindent\textbf{Fine-tuning for VLMs.}
Inspired by the prior work in NLP, recent researches focus on developing efficient fine-tuning methods for VLMs on downstream tasks.
One type of such method is prompt tuning, which has been explored in several recent works~\cite{zhou2022learning, lu2022prompt, chen2022prompt}.
CoOp~\cite{zhou2022learning} proposes a prompt learning method that optimizes a class-agnostic prompt template in the continuous token embedding space through back-propagation on few-shot datasets.
ProDA~\cite{lu2022prompt} attempts to learn a collection of continuous prompts to capture the variational visual representation.
PLOT~\cite{chen2022prompt} proposes to apply optimal transport to match the learnable prompts with different areas of the images.
Another type of fine-tuning method is adapters~\cite{gao2021clip, zhang2022tip}.
CLIP-Adapter~\cite{gao2021clip} proposes to add a lightweight MLP following the last vision layer and mix the output feature with the original zero-shot feature via a residual connection.
Tip-Adapter~\cite{zhang2022tip} further improves CLIP-Adapter~\cite{gao2021clip} by replacing the lightweight MLP with a linear layer, whose weights are comprised of the labeled visual embeddings, acting as visual prototypes of the concepts.
This not only inherits the training-free advantage of zero-shot CLIP~\cite{radford2021learning} but also performs comparably to those training-required approaches.

While these methods have achieved significant improvements on downstream datasets, they require data for all classes when fine-tuning.
When dealing with new unseen classes, they either become invalid~\cite{zhang2022tip} or their performance drops dramatically~\cite{zhou2022learning}. 
However, some classes are difficult to collect data for because of their rareness, such as new species or concepts.  
As a result, many categories may be devoid of any relevant data.
To address this, previous methods have attempted to learn more robust prompts.
CoCoOp~\cite{zhou2022conditional} improves new class performance by learning an instance-specific continuous prompt conditioned on the input image.
With image information, the prompts are easily transferred to recognize new class samples.
VPT~\cite{derakhshani2022variational} proposes to learn the distribution of instance-specific prompts via variational inference.
During inference, VPT ensembles several prompts sampled from the distribution for the classification.
In contrast to the previous methods~\cite{zhou2022conditional, derakhshani2022variational}, we propose to synthesize features for those unseen categories.
With features for all classes, we can utilize off-the-shelf methods to fine-tune CLIP.

\noindent\textbf{Generalized Zero-Shot Learning.}
Generalized zero-shot learning (GZSL) is a relevant research field with similar objectives to our work.
Specifically, GZSL focuses on training a classifier that can recognize both seen and unseen object classes, where the latter is absent from the training set.
To accomplish this, GZSL leverages auxiliary semantic information such as expert annotated attributes or text descriptions~\cite{mikolov2013distributed} for both seen and unseen classes.
Embedding-based GZSL methods aim to learn a visual-to-semantic mapping for visual-semantic interaction by mapping visual features into the semantic space~\cite{xu2020attribute, xie2019attentive}.
However, a major drawback of these methods is their bias towards seen classes, as they only learn from seen data.
As a solution, generative-based GZSL methods have been introduced to learn semantic-to-visual mapping to generate visual features of unseen classes~\cite{xian2018feature, xian2019f, li2019leveraging, narayan2020latent} for data augmentation.
Currently, the generative methods are typically based on variational autoencoders (VAEs)~\cite{kingma2013auto, xian2019f}, generative adversarial networks (GANs)~\cite{xian2018feature, xian2019f, li2019leveraging,felix2018multi}, and generative flows~\cite{shen2020invertible}. 
Despite their promising results, these generative-based methods require training on a large seen dataset to learn semantic-visual mapping and expertly annotated attributes or text descriptions for all classes, which can be labor-intensive.
In our work, we aim to imitate GZSL by learning to synthesize samples for new classes.
However, with limited labeled data in the training set and coarse semantic vectors for each class through prompting like ``a photo of a \{class\}", these GZSL generative methods fail to synthesize valid new samples for new classes. 

\section{Method}
\label{sec:method}
\begin{figure*}[htbp]
    \centering
    \includegraphics[width=0.9\textwidth]{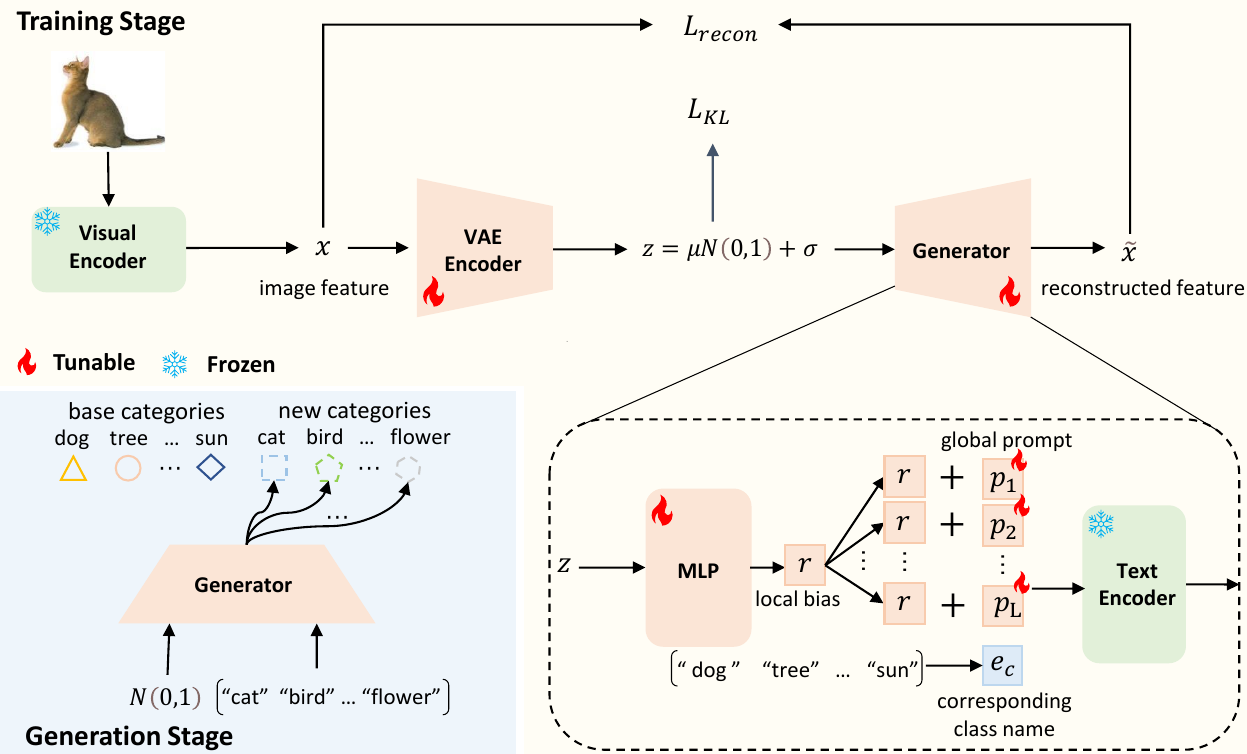}
    \caption{
    The proposed model architecture is built upon the VAE framework, comprising a VAE encoder and a generator.
    In the training stage, we extract the image feature with CLIP visual encoder and the VAE encoder encodes it into a latent code $z$, which is subsequently confined to a prior distribution.  
    Following this, the generator reconstructs the input feature by utilizing the encoded information.
    Notably, a novel CLIP-based generator is introduced, which comprises two subnetworks: a lightweight MLP and a frozen CLIP text encoder.
    The MLP transforms the latent code $z$ into a local bias, which is subsequently added to global learnable prompt vectors to construct the final prompts.
    The prompts, together with the class names, are then input into the frozen text encoder to obtain the reconstructed feature.
    During the generating stage, we sample the latent code from the prior distribution and then use it with the new class name to synthesize the corresponding features.
    Finally, we fine-tune CLIP using off-the-shelf methods with the base class and synthetic new class features.
    }
    \label{fig:architecture}
\end{figure*}

\subsection{Background}
Contrastive Language-Image Pretraining, known as CLIP~\cite{radford2021learning}, is a method developed for aligning the representations of images and their corresponding captions, which has gained considerable attention in recent years.
CLIP consists of two encoder modules: a visual encoder $\mathcal{I}(\bm{x})$ and a language encoder $\mathcal{T}(t)$, which encode images and text descriptions, respectively, into a shared $d$-dimensional space. 
The visual encoder can be ViT~\cite{dosovitskiyimage} or ResNet~\cite{he2016deep}, while the language encoder is a Transformer~\cite{vaswani2017attention}.
Both encoders are trained jointly using a contrastive loss applied to a large dataset of paired images and captions.
Once trained, CLIP can be used for zero-shot classification of downstream tasks. 
To perform $C$-class image classification, category descriptions $\{t_c\}_{c=1}^C$ are generated through prompting, such as ``a photo of a \{class\}". Then, the classification probability of the input image $\bm{x}$ is computed as follows: 
\begin{equation}
    p(y|\bm{x}) = \frac{\exp(\cos(\mathcal{I}(\bm{x}), \mathcal{T}(t_y)) / \tau)}{\sum_{c=1}^C\exp(\cos(\mathcal{I}(\bm{x}), \mathcal{T}(t_c)) / \tau)},
\end{equation}
where $\tau$ denotes the temperature, $\cos(\cdot, \cdot)$ is the cosine similarity function, and $y$ is the target class.

\subsection{Synthesized Prompts}
In this paper, we aim to improve the performance of CLIP on both base and new categories, i.e., categories with and without available data, while maintaining data efficiency as previous methods.
To achieve this goal, a novel generative approach named \textbf{S}ynt\textbf{H}es\textbf{I}zed \textbf{P}rompts~(\textbf{SHIP}) is proposed, which involves three stages.
First, we follow variational autoencoders to introduce a generator that reconstructs the visual features by inputting the synthesized prompts and the corresponding class names to the language encoder of CLIP.
Subsequently, we obtain the synthesized features for new categories by providing the class names.
Finally, we combine the labeled base class features with the synthesized new class features and employ existing fine-tuning methods, such as CoOp~\cite{zhou2022learning} and Tip-Adapter~\cite{zhang2022tip}, to fine-tune CLIP, which thus enhances its performance on both base and new classes.

The architecture of the generative model is illustrated in Figure~\ref{fig:architecture}.
To maintain the data efficiency, we opt to employ the Variational Autoencoder (VAE)~\cite{kingma2013auto} for training our generator instead of Generative Adversarial Networks (GANs)~\cite{goodfellow2020generative}.
The reason is that it is difficult to train an effective discriminator for GANs with limited labeled data~\cite{karras2020training}.
As depicted in Figure~\ref{fig:architecture}, the VAE architecture comprises an encoder $E(x)$ and a generator $G(z, c)$.
First, we leverage the fixed CLIP visual encoder to extract the feature of the input image, i.e., ${x} = \mathcal{I}(img)$.
Subsequently, the VAE encoder $E(x)$ encodes the feature ${x}$ into a latent code ${z}$, and the generator $G(z, c)$ reconstructs the feature $x$ using the latent code $z$ and the corresponding class name $c$.
The optimization of both $E$ and $G$ is achieved via the evidence-lower bound given by the equation as follows:
\begin{equation}
\begin{aligned}
    L & = L_{recon} + L_{KL} \\
      & = \mathbb{E}[-\log G(z, c)] + KL(E(x)\|p(z|c)), 
\end{aligned}
\end{equation}
where $KL$ represents the Kullback-Leibler divergence, $p(z|c)$ is a prior distribution that is assumed to be $\mathcal{N}(0, 1)$, and $-\log G(z, c)$ denotes the reconstruction loss.

To further utilize the pretrained knowledge of CLIP, we propose a CLIP-based generator. 
Notably, the pretrained CLIP has learned aligned vision and language representations, allowing us to reconstruct input features from the language encoder $\mathcal{T}$.
Since having been trained on a large-scale dataset, the reconstructed features obtained from the pretrained language model $\mathcal{T}$ are expected to be of higher quality than those generated by a new generator trained from scratch on the few-shot base dataset.
Drawing inspiration from previous prompt learning methods~\cite{zhou2022learning, zhou2022conditional, lu2022prompt}, we generate instance-specific prompts, instead of generating the features directly.
Specifically, given the latent code $z$, we generate instance-specific prompts as follows:
\begin{equation}
    \label{eq1}
    \bm{p}(z) = [\bm{p}_1 + \bm{r}, \bm{p}_2 + \bm{r}, ..., \bm{p}_L + \bm{r}], 
\end{equation}
where the local bias $\bm{r}$ is obtained through a two-layer fully-connected network, i.e., $\bm{r}=MLP(z)$, that embeds the latent code $z$ into the token embedding space, and $L$ is the length of prompts.
As in Eq.~(\ref{eq1}), our prompts consist of two components: a global fixed set of learnable prompts $\{\bm{p}_i, i=1,2,..., L\}$, which are randomly initialized, capturing the global information of the input features and a local bias $\bm{r}$ that encodes the instance-specific information of the input feature into the prompts.
By combining the prompts and the token embedding of the corresponding class name, we obtain the reconstructed features as follows:
\begin{equation}
    \bm{\Tilde{x}} = \mathcal{T}(t), \quad t = \{\bm{p}(z), \bm{e_c}\},
\end{equation}
where $\mathcal{T}$ is the frozen language encoder, and $\bm{e_c}$ is the token embedding of the corresponding class names. 

During the training stage, we maintain the CLIP frozen and only optimize the encoder $E$, the lightweight $MLP$, and the global prompts $\bm{p} = [\bm{p_1}, \bm{p_2}, ..., \bm{p_L}]$.

\subsection{Fine-tuning CLIP}
Following the training stage, the generator is employed to synthesize features for new classes.
Specifically, given the class name $c$ of a new class and the noise $z$ sampled from the prior distribution, the generator $G(z, c)$ is utilized to generate the corresponding features. 
This process is repeated for each new class, resulting in a new synthetic dataset.
When combined with the labeled base dataset, a complete dataset for all classes is obtained.
Consequently, off-the-shelf methods~\cite{zhou2022learning, gao2021clip, zhang2022tip, chen2022prompt} can be employed to fine-tune CLIP, which is expected to perform better on new classes in comparison to its previous counterparts.

\section{Experiments}

\subsection{Setup}
We evaluate our method for three different tasks: base-to-new generalization, cross-dataset transfer, and generalized zero-shot classification. 
For the base-to-new generalization and cross-dataset transfer tasks, we follow the same experimental setting as CoCoOp~\cite{zhou2022conditional}. 
It uses a total of 11 diverse image classification datasets, i.e., ImageNet~\cite{deng2009imagenet} and Caltech101~\cite{fei2004learning} for generic object recognition, OxfordPets~\cite{parkhi2012cats}, StanfordCars~\cite{krause20133d}, Flowers102~\cite{nilsback2008automated}, Food101~\cite{bossard2014food} and FGVCAircraft~\cite{maji2013fine} for fine-grained image recognition, EuroSAT~\cite{helber2019eurosat} for satellite image classification, UCF101~\cite{soomro2012ucf101} for action classification, DTD~\cite{cimpoi2014describing} for texture classification, and SUN397~\cite{xiao2010sun} for scene recognition. 
For generalized zero-shot classification tasks, we follow the same setting as \cite{xian2018zero}, and we conduct the experiments on three standard zero-shot recognition datasets: Caltech-UCSD-Birds~\cite{welinder2010caltech} (CUB), Oxford Flowers~\cite{nilsback2008automated} (FLO), and Animals with Attributes2~\cite{xian2018zero} (AWA2), containing 200, 102, and 50 categories, respectively. For a fair comparison, we use the same data splits and evaluation protocols as proposed in ~\cite{xian2018zero}.
\\
\noindent
\textbf{Implementation details.}
Our proposed method is comprised of three sub-networks: a VAE encoder, a lightweight MLP, and a pretrained CLIP. 
The VAE encoder and the MLP are implemented as two-layer fully-connected networks with 4,096 hidden units and ReLU activation. 
And we employ ViT-B/16~\cite{dosovitskiyimage} and transformer~\cite{vaswani2017attention} as the vision and language encoders of CLIP, which are initialized with CLIP's pretrained weights and kept frozen during training. 
The dimensions of the latent code $z$ are set to be equal to the dimension of token embedding.
We fix the length of the learnable global context vectors to 4 and initialize them with Gaussian noise.
The features are normalized to a unit sphere, as proposed in CLIP~\cite{radford2021learning}.
And we utilize MSE as the reconstruction loss of the VAE.
All the networks are trained using the AdamW optimizer with a learning rate of 0.001. 
During the fine-tuning of CLIP, since we utilize off-the-shelf methods, we follow the same settings as those proposed in their papers~\cite{zhou2022conditional, zhou2022learning, zhang2022tip, gao2021clip}.
We randomly synthesize a batch of new class features and combine them with the original batch to form a new batch during training.
We conduct all experiments on a single NVIDIA GeForce RTX 3090, except for the ImageNet dataset, which is conducted on an NVIDIA A100.

\setlength{\tabcolsep}{10.5pt}
\begin{table*}[htbp]
  \centering
  \caption{
  \textbf{Base-to-new generalization.}
  Our proposed model is trained on a few-shot training set (base) and then evaluated on both base and new classes. 
  +SHIP denotes we add our method to previous off-the-shelf methods.
  The result of Tip-Adapter~\cite{zhang2022tip} is not included in the table due to its inability to test on new classes.
  The average accuracy of the base and new classes is represented by the terms \textbf{Base} and \textbf{New}, respectively, while their harmonic mean is denoted as \textbf{H}.
  The best results are presented in bold.
  }
  \resizebox{1.0\textwidth}{!}{
    \begin{tabular}{ll|rrr|rrr|rrr|rrr}
    \toprule
    \multicolumn{2}{c|}{\multirow{2}[2]{*}{}} & \multicolumn{3}{c|}{\textbf{Average}} & \multicolumn{3}{c|}{\textbf{ImageNet~\cite{deng2009imagenet}}} & \multicolumn{3}{c|}{\textbf{Caltech101~\cite{fei2004learning}}} & \multicolumn{3}{c}{\textbf{OxfordPets~\cite{parkhi2012cats}}} \\
    \multicolumn{2}{c|}{} & \multicolumn{1}{c}{\textbf{Base}} & \multicolumn{1}{c}{\textbf{New}} & \multicolumn{1}{c|}{\textbf{H}} & \multicolumn{1}{c}{\textbf{Base}} & \multicolumn{1}{c}{\textbf{New}} & \multicolumn{1}{c|}{\textbf{H}} & \multicolumn{1}{c}{\textbf{Base}} & \multicolumn{1}{c}{\textbf{New}} & \multicolumn{1}{c|}{\textbf{H}} & \multicolumn{1}{c}{\textbf{Base}} & \multicolumn{1}{c}{\textbf{New}} & \multicolumn{1}{c}{\textbf{H}} \\
    \midrule
    \multicolumn{2}{l|}{CLIP~\cite{radford2021learning}} & 69.34  & 74.22  & 71.70  & 72.43  & 68.14  & 70.22  & 96.84  & 94.00  & 95.40  & 91.17  & 97.26  & 94.12  \\
    \multicolumn{2}{l|}{CoOp~\cite{zhou2022learning}} & 82.69  & 63.22  & 71.66  & 76.47  & 67.88  & 71.92  & 98.00  & 89.81  & 93.73  & 93.67  & 95.29  & 94.47  \\
    \multicolumn{2}{l|}{CoCoOp~\cite{zhou2022conditional}} & 80.47  & 71.69  & 75.83  & 75.98  & 70.43  & 73.10  & 97.96  & 93.81  & 95.84  & 95.20  & 97.69  & 96.43  \\
    \multicolumn{2}{l|}{ProDA~\cite{lu2022prompt}} & 81.56  & 72.30  & 76.65  & 75.40  & 70.23  & 72.72  & 98.27  & 93.23  & 95.68  & \textbf{95.43} & 97.83  & \textbf{96.62} \\
    \multicolumn{2}{l|}{CLIP-Adapter~\cite{lu2022prompt}} & 83.05  & 65.20  & 73.05  & 75.74  & 68.21  & 71.78  & 98.13  & 92.19  & 95.39  & 91.55  & 90.10  & 90.82  \\
    \multicolumn{2}{l|}{CoOp + VPT~\cite{derakhshani2022variational}} & 71.98  & 74.76  & 73.34  & 74.73  & \textbf{70.60} & 72.60  & 95.47  & 93.80  & 94.62  & 90.77  & 97.83  & 94.16  \\
    \multicolumn{2}{l|}{CoOp + SHIP} & 80.03  & 73.69  & 76.73  & 75.87  & 69.95  & 72.79  & 97.55  & \textbf{95.20} & 96.36 & 95.37  & \textbf{97.87} & 96.61  \\  
    \multicolumn{2}{l|}{CLIP-Adapter + SHIP} & 83.14  & 67.77  & 74.67  & 76.00  & 69.32  & 72.51  & 97.68  & 95.09  & \textbf{96.37} & 92.19  & 93.85  & 93.01  \\ 
    \multicolumn{2}{l|}{Tip-Adapter + SHIP} & \textbf{83.80} & \textbf{76.42} & \textbf{79.94} & \textbf{77.53} & 70.26  & \textbf{73.71} & \textbf{98.32} & 94.43  & 96.34  & 94.95  & 97.09  & 96.01  \\ 
    \bottomrule
    \end{tabular}%
  }
    \vspace{0.1cm}\\
   \resizebox{1.0\textwidth}{!}{
    \begin{tabular}{ll|rrr|rrr|rrr|rrr}
    \toprule
    \multicolumn{2}{c|}{\multirow{2}[2]{*}{}} & \multicolumn{3}{c|}{\textbf{StanfordCars~\cite{krause20133d}}} & \multicolumn{3}{c|}{\textbf{Flowers102~\cite{nilsback2008automated}}} & \multicolumn{3}{c|}{\textbf{Food101~\cite{bossard2014food}}} & \multicolumn{3}{c}{\textbf{FGVCAircraft~\cite{maji2013fine}}} \\
    \multicolumn{2}{c|}{} & \multicolumn{1}{c}{\textbf{Base}} & \multicolumn{1}{c}{\textbf{New}} & \multicolumn{1}{c|}{\textbf{H}} & \multicolumn{1}{c}{\textbf{Base}} & \multicolumn{1}{c}{\textbf{New}} & \multicolumn{1}{c|}{\textbf{H}} & \multicolumn{1}{c}{\textbf{Base}} & \multicolumn{1}{c}{\textbf{New}} & \multicolumn{1}{c|}{\textbf{H}} & \multicolumn{1}{c}{\textbf{Base}} & \multicolumn{1}{c}{\textbf{New}} & \multicolumn{1}{c}{\textbf{H}} \\
    \midrule
    \multicolumn{2}{l|}{CLIP~\cite{radford2021learning}} & 63.37  & 74.89  & 68.65  & 72.08  & 77.80  & 74.83  & 90.10  & 91.22  & 90.66  & 27.19  & \textbf{36.29} & 31.09  \\
    \multicolumn{2}{l|}{CoOp~\cite{zhou2022learning}} & 78.12  & 60.40  & 68.13  & 97.60  & 59.67  & 74.06  & 88.33  & 82.26  & 85.19  & 40.44  & 22.30  & 28.75  \\
    \multicolumn{2}{l|}{CoCoOp~\cite{zhou2022conditional}} & 70.49  & 73.59  & 72.01  & 94.87  & 71.75  & 81.71  & \textbf{90.70} & 91.29  & 90.99  & 33.41  & 23.71  & 27.74  \\
    \multicolumn{2}{l|}{ProDA~\cite{lu2022prompt}} & 74.70  & 71.20  & 72.91  & 97.70  & 68.68  & 80.66  & 90.30  & 88.57  & 89.43  & 36.90  & 34.13  & 35.46  \\
    \multicolumn{2}{l|}{CLIP-Adapter~\cite{gao2021clip}} & 79.16  & 59.49  & 67.93  & \textbf{98.29} & 64.68  & 78.02  & 88.24  & 88.33  & 88.29  & 42.14  & 25.67  & 31.91  \\
    \multicolumn{2}{l|}{CoOp + VPT~\cite{derakhshani2022variational}} & 65.27  & \textbf{75.97} & 70.21  & 72.97  & 75.90  & 74.40  & 90.37  & \textbf{91.67} & 91.01  & 29.57  & 33.80  & 31.54  \\
    \multicolumn{2}{l|}{CoOp + SHIP} & 68.57  & 73.90  & 71.14  & 94.02  & 74.40  & 83.06  & 90.54  & 91.03  & 90.78  & 34.27  & 32.33  & 33.28  \\
    \multicolumn{2}{l|}{CLIP-Adapter + SHIP} & 78.51  & 62.52  & 69.61  & 98.20  & 65.89  & 78.86  & 88.63  & 87.07  & 87.84  & 42.26  & 30.05  & 35.13  \\
    \multicolumn{2}{l|}{Tip-Adapter + SHIP} & \textbf{79.91} & 74.62  & \textbf{77.18} & 95.35  & \textbf{77.87} & \textbf{85.73} & 90.63  & 91.51  & \textbf{91.07} & \textbf{42.62} & 35.93  & \textbf{38.99} \\
    \bottomrule
    \end{tabular}%
  }
    \vspace{0.1cm}\\
    \resizebox{1.0\textwidth}{!}{
    \begin{tabular}{ll|rrr|rrr|rrr|rrr}
    \toprule
    \multicolumn{2}{c|}{\multirow{2}[2]{*}{}} & \multicolumn{3}{c|}{\textbf{SUN397~\cite{xiao2010sun}}} & \multicolumn{3}{c|}{\textbf{DTD~\cite{cimpoi2014describing}}} & \multicolumn{3}{c|}{\textbf{EuroSAT~\cite{helber2019eurosat}}} & \multicolumn{3}{c}{\textbf{UCF101~\cite{soomro2012ucf101}}} \\
    \multicolumn{2}{c|}{} & \multicolumn{1}{c}{\textbf{Base}} & \multicolumn{1}{c}{\textbf{New}} & \multicolumn{1}{c|}{\textbf{H}} & \multicolumn{1}{c}{\textbf{Base}} & \multicolumn{1}{c}{\textbf{New}} & \multicolumn{1}{c|}{\textbf{H}} & \multicolumn{1}{c}{\textbf{Base}} & \multicolumn{1}{c}{\textbf{New}} & \multicolumn{1}{c|}{\textbf{H}} & \multicolumn{1}{c}{\textbf{Base}} & \multicolumn{1}{c}{\textbf{New}} & \multicolumn{1}{c}{\textbf{H}} \\
    \midrule
    \multicolumn{2}{l|}{CLIP~\cite{radford2021learning}} & 69.36  & 75.35  & 72.23  & 53.24  & 59.90  & 56.37  & 56.48  & 64.05  & 60.03  & 70.53  & 77.50  & 73.85  \\
    \multicolumn{2}{l|}{CoOp~\cite{zhou2022learning}} & 80.60  & 65.89  & 72.51  & 79.44  & 41.18  & 54.24  & 92.19  & 54.74  & 68.69  & 84.69  & 56.05  & 67.46  \\
    \multicolumn{2}{l|}{CoCoOp~\cite{zhou2022conditional}} & 79.74  & 76.86  & 78.27  & 77.01  & 56.00  & 64.85  & 87.49  & 60.04  & 71.21  & 82.33  & 73.45  & 77.64  \\
    \multicolumn{2}{l|}{ProDA~\cite{lu2022prompt}} & 78.67  & 76.93  & 77.79  & 80.67  & 56.48  & 66.44  & 83.90  & 66.00  & 73.88  & 85.23  & 71.97  & 78.04  \\
    \multicolumn{2}{l|}{CLIP-Adapter~\cite{gao2021clip}} & 79.44  & 66.81  & 72.58  & \textbf{81.94} & 39.49  & 53.30  & \textbf{93.45} & 54.41  & 68.78  & 85.42  & 67.77  & 75.58  \\
    \multicolumn{2}{l|}{CoOp + VPT~\cite{derakhshani2022variational}} & 73.77  & \textbf{77.90} & 75.77  & 57.67  & 58.70  & 58.18  & 67.97  & 71.63  & 69.75  & 73.23  & 74.63  & 73.92  \\
    \multicolumn{2}{l|}{CoOp + SHIP} & 79.54  & 75.27  & 77.35  & 74.88  & 56.88  & 64.65  & 88.62  & 66.87  & 76.22  & 81.08  & 76.85  & 78.91  \\
    \multicolumn{2}{l|}{CLIP-Adapter + SHIP} & 79.86  & 66.52  & 72.58  & 81.60  & 46.38  & 59.14  & 93.05  & 57.15  & 70.81  & \textbf{86.61} & 71.61  & 78.40  \\
    \multicolumn{2}{l|}{Tip-Adapter + SHIP} & \textbf{81.32} & 77.64  & \textbf{79.43} & 81.83  & \textbf{61.47} & \textbf{70.21} & 93.38  & \textbf{81.67} & \textbf{87.13} & 85.99  & \textbf{78.10} & \textbf{81.85} \\
    \bottomrule
    \end{tabular}%
  }
  \label{tab:b2n}%
\end{table*}%

\subsection{Results}
\subsubsection{Base-to-new generalization}
\noindent\textbf{Setup.} 
Following CoCoOp~\cite{zhou2022conditional}, we partition each dataset into two equal non-overlapping subsets: the base classes and the new classes. 
 Subsequently, we randomly extract a few-shot training set from base classes, while preserving the original test set for evaluation purposes.
Specifically, we perform training on the base classes with a mere 16 samples per class and evaluate the trained model on both the base and new classes.
To evaluate the model's performance, we compute the average accuracy of both the base and new classes, as well as their harmonic mean~\cite{zhou2022conditional} ($H = 2\times base\times new / (base + new)$).

\noindent\textbf{Results.} 
We choose CLIP~\cite{radford2021learning}, CoOp~\cite{zhou2022learning}, CoCoOp~\cite{zhou2022conditional}, CLIP-Adapter~\cite{gao2021clip}, Tip-Adapter~\cite{zhang2022tip}, VPT~\cite{derakhshani2022variational}, and ProDA~\cite{lu2022prompt} as our baseline. 
The result of Tip-Adapter~\cite{zhang2022tip} is not included in the table due to its inability to test on new classes.
Results from Table~\ref{tab:b2n} show that the previous fine-tuning methods significantly degrade the performance of CLIP on new classes.
Specifically, CoOp~\cite{zhou2022learning} reduces the accuracy of new classes by an average of 11\% across 11 datasets.
Tip-Adapter~\cite{zhang2022tip} is even worse as it fails to recognize new categories outside the training set. 
It is noteworthy that all previous methods, except VPT~\cite{derakhshani2022variational}, harm the CLIP performance on new classes.
However, VPT~\cite{derakhshani2022variational} achieves this by reducing the base class accuracy by 10.7\%. 

As shown in Table~\ref{tab:b2n}, we add our generative prompt tuning method to three baseline methods: CoOp~\cite{zhou2022learning}, CLIP-Adapter~\cite{gao2021clip}, and Tip-Adapter~\cite{zhang2022tip}.
By adding our method, CoOp + SHIP outperforms CoOp~\cite{zhou2022learning} by 10.47\% and 5.07\% on the new classes and harmonic mean, respectively, while only sacrificing 2.66\% on the base classes.
The incorporation of generative prompt tuning into CLIP-Adapter~\cite{gao2021clip} results in a 2.57\% and 1.62\% improvement in performance on the new classes and harmonic mean, respectively, without affecting the performance of the base classes.
Notably, augmenting Tip-Adapter~\cite{zhang2022tip} with our proposed generative prompt tuning method not only expands its recognition ability to new classes but also achieves almost the best results compared to all the baseline methods.
Specifically, Tip-Adapter + SHIP achieves a 14.46\% improvement on the base classes, 2.20\% on the new classes, and 8.24\% on the harmonic mean on average across all datasets compared to zero-shot CLIP.
Moreover, it obtains the highest harmonic mean on nine of the eleven datasets, except for Caltech101~\cite{fei2004learning} and OxfordPets~\cite{parkhi2012cats}, where the performance has already reached a high level ($>95\%$), thus limiting the potential for improvement.

\setlength{\tabcolsep}{10.0pt}
\begin{table*}[htbp]
  \centering
  \caption{
  \textbf{Cross dataset transfer learning.} 
  The methods are trained on a source dataset (ImageNet) and subsequently evaluated on target datasets. 
  We report the average accuracy of the target datasets.
  To quantify the performance gains of our method, we compute the difference between the results obtained using our approach ($CoOp + SHIP$) and the baseline approach ($CoOp$).
  }
  \resizebox{1.\textwidth}{!}{
    \begin{tabular}{ll|rrrrrrrrrrr}
    \toprule
    \multicolumn{2}{c|}{\textbf{Method}} & {\begin{turn}{70}\textbf{Caltech101~\cite{fei2004learning}}\end{turn}} & {\begin{turn}{70}\textbf{OxfordPets~\cite{parkhi2012cats}}\end{turn}} & {\begin{turn}{70}\textbf{StanfordCars~\cite{krause20133d}}\end{turn}} & {\begin{turn}{70}\textbf{Flowers102~\cite{nilsback2008automated}}\end{turn}} & {\begin{turn}{70}\textbf{Food101~\cite{bossard2014food}}\end{turn}} & {\begin{turn}{70}\textbf{FGVC~\cite{maji2013fine}}\end{turn}} & {\begin{turn}{70}\textbf{SUN397~\cite{xiao2010sun}}\end{turn}} & {\begin{turn}{70}\textbf{DTD~\cite{cimpoi2014describing}}\end{turn}} & {\begin{turn}{70}\textbf{EuroSAT~\cite{helber2019eurosat}}\end{turn}} & {\begin{turn}{70}\textbf{UCF101~\cite{soomro2012ucf101}}\end{turn}} &{\begin{turn}{70}\textbf{\textit{Average}}\end{turn}} \\
    \midrule
    \multicolumn{2}{l|}{CLIP~\cite{radford2021learning}} & 92.94  & 89.21  & 65.32  & \textbf{71.34}  & 86.06  & \textbf{24.72}  & 62.50  & 44.39  & 47.60  & 66.75  & 65.08  \\
    \multicolumn{2}{l|}{CoOp~\cite{zhou2022learning}} & 93.70  & 89.14  & 64.51  & 68.71  & 85.30  & 18.47  & 64.15  & 41.92  & 46.39  & 66.55  & 63.88  \\
    \multicolumn{2}{l|}{CoOp + SHIP} & \textbf{94.04} & \textbf{90.38}  & \textbf{65.55}  & 69.67  & \textbf{86.40}  & 21.90  & \textbf{66.62}  & \textbf{45.69}  & \textbf{48.17}  & \textbf{68.52}  & \textbf{65.69}  \\
    \midrule
    \multicolumn{2}{l|}{$\Delta$} & \textcolor[rgb]{0,0,1}{+0.34}  & \textcolor[rgb]{0,0,1}{+1.24}  & \textcolor[rgb]{0,0,1}{+1.04}  & \textcolor[rgb]{0,0,1}{+0.96}  & \textcolor[rgb]{0,0,1}{+1.10}  & \textcolor[rgb]{0,0,1}{+3.43}  & \textcolor[rgb]{0,0,1}{+2.47}  & \textcolor[rgb]{0,0,1}{+3.77}  & \textcolor[rgb]{0,0,1}{+1.78}  & \textcolor[rgb]{0,0,1}{+1.97}  & \textcolor[rgb]{0,0,1}{+1.81}  \\
    \bottomrule
    \end{tabular}%
    }
  \label{tab:cross}%
\end{table*}%

\subsubsection{Cross-dataset transfer learning}
\noindent\textbf{Setup.} 
Following CoCoOp~\cite{zhou2022conditional}, we present an evaluation of our method's cross-dataset transfer performance.
Specifically, we examine the effectiveness of our approach on ten different target datasets following training on the source dataset (ImageNet~\cite{deng2009imagenet}).
To simulate more realistic scenarios, we train our generative model and CoOp~\cite{zhou2022learning} on 16-shot ImageNet, utilizing all 1,000 available classes.
Subsequently, using the generative model, we generate features for all classes in the target dataset and fine-tune CoOp~\cite{zhou2022learning} with the synthesized data.
We report the average accuracy of these datasets for a fair comparison.

\noindent\textbf{Results.}
We report the performance of the proposed CoOp + SHIP compared to the CoOp~\cite{zhou2022learning} and CLIP~\cite{radford2021learning} in ten target datasets.
The results are shown in Table~\ref{tab:cross}, indicating an improvement range of 0.34\% to 3.77\%, with an average improvement of 1.81\%.
Notably, the CoOp + SHIP outperformed the baselines in eight out of ten datasets, with exceptions in Flowers102~\cite{nilsback2008automated} and FGVCAircraft~\cite{maji2013fine} datasets.
The reason for this observation is that Flowers102~\cite{nilsback2008automated} and FGVCAircraft~\cite{maji2013fine} are fine-grained datasets that pose a challenge for the generator to synthesize in-distribution and non-trivial features.

\setlength{\tabcolsep}{8pt}
\begin{table*}[htbp]
  \centering
  \caption{
  \textbf{Generalized zero-shot learning.}
  Models are trained on seen class data and evaluated on the mixture of seen and unseen test datasets.
  We evaluate on three datasets: CUB~\cite{welinder2010caltech}, AWA2~\cite{xian2018zero}, and FLO~\cite{nilsback2008automated}.
  Results are reported in terms of average \textit{top-1} accuracy of unseen and seen classes, together with their harmonic mean~(H).
  }
    \resizebox{0.9\textwidth}{!}{
    \begin{tabular}{cll|ccc|ccc|ccc}
    \toprule
    \multicolumn{3}{c|}{\multirow{2}[2]{*}{\textbf{Method}}} &\multicolumn{3}{c|}{CUB~\cite{welinder2010caltech}} & \multicolumn{3}{c|}{AWA2~\cite{xian2018zero}} & \multicolumn{3}{c}{FLO~\cite{nilsback2008automated}} \\
    
    \multicolumn{3}{c|}{} &\multicolumn{1}{c}{\textbf{Unseen}} & \multicolumn{1}{c}{\textbf{Seen}} & \multicolumn{1}{c|}{\textbf{H}} & \multicolumn{1}{c}{\textbf{Unseen}} & \multicolumn{1}{c}{\textbf{Seen}} & \multicolumn{1}{c|}{\textbf{H}} & \multicolumn{1}{c}{\textbf{Unseen}} & \multicolumn{1}{c}{\textbf{Seen}} & \multicolumn{1}{c}{\textbf{H}} \\
    \midrule
    \multirow{9}[2]{*}{\begin{sideways}Resnet-101\end{sideways}} 
          & \multicolumn{2}{l|}{f-CLSWGAN~\cite{xian2018feature}} & 3.7  & 57.7  & 49.7  & 57.9  & 61.4  & 59.6  & 59.0  & 73.8  & 65.6  \\
          & \multicolumn{2}{l|}{Cycle-WGAN~\cite{felix2018multi}} & 47.9  & 59.3  & 53.0  & 59.6  & 63.4  & 59.8  & 61.6  & 69.2  & 65.2  \\
          & \multicolumn{2}{l|}{LisGAN~\cite{li2019leveraging}} & 46.5  & 57.9  & 51.6  & 52.6  & 76.3  & 62.3  & 57.7  & 83.8  & 68.3  \\
          & \multicolumn{2}{l|}{TCN~\cite{jiang2019transferable}} & 52.6  & 52.0  & 52.3  & 61.2  & 65.8  & 63.4  & -     & -     & - \\
          & \multicolumn{2}{l|}{f-VAEGAN~\cite{xian2019f}} & 48.4  & 60.1  & 53.6  & 57.6  & 70.6  & 63.5  & 56.8  & 74.9  & 64.6  \\
          & \multicolumn{2}{l|}{TF-VAEGAN~\cite{narayan2020latent}} & 52.8  & 64.7  & 58.1  & 59.8  & 75.1  & 66.6  & 62.5  & 84.1  & 71.7  \\
          & \multicolumn{2}{l|}{GCM-CF~\cite{yue2021counterfactual}} & \textbf{61.0}  & 59.7  & 60.3  & 60.4  & 75.1  & 67.0  & -  & -  & -  \\
          & \multicolumn{2}{l|}{HSVA~\cite{chen2021hsva}} & 52.7  & 58.3  & 55.3  & 56.7  & 79.8  & 66.3  & -  & -  & -  \\
          & \multicolumn{2}{l|}{MSDN~\cite{chen2022msdn}} & {68.7}  & {67.5}  & \textbf{68.1}  & 62.0  & 74.5  & 67.7  & -     & -     & - \\
    \midrule
    \multirow{5}[2]{*}{\begin{sideways}CLIP\end{sideways}} & \multicolumn{2}{l|}{CLIP~\cite{radford2021learning}} & 55.2  & 54.8  & 55.0  & \textbf{88.3}  & 93.1  & \textbf{90.6}  & 65.6  & 67.9  & 66.7  \\
          & \multicolumn{2}{l|}{CoOp~\cite{zhou2022learning}} & 49.2  & 63.8  & 55.6  & 72.7  & {95.3}  & 82.5  & 52.2  & {85.8}  & 64.9  \\
          & \multicolumn{2}{l|}{TF-VAEGAN~\cite{narayan2020latent}} & 21.1  & \textbf{84.4}  & 34.0  & 43.7  & \textbf{96.3}  & 60.1  & 37.4  & 97.2  & 54.0  \\
          & \multicolumn{2}{l|}{f-VAEGAN~\cite{xian2019f}} & 22.5  & 82.2  & 35.3  & 61.2  & 95.9  & 74.7  & 11.1  & \textbf{97.6}  & 20.0  \\
          & \multicolumn{2}{l|}{CoOp + SHIP} & 55.3  & 58.9  & 57.1  & 84.1  & 94.4  & 89.0  & \textbf{69.0}  & 76.3  & \textbf{72.4}  \\
    \bottomrule
    \end{tabular}%
    }
  \label{tab:gzsl}%
\end{table*}%

\subsubsection{Generalized zero-shot learning}
\noindent\textbf{Setup.} 
We follow the same data split and evaluation metrics as in~\cite{xian2018zero}.
To ensure fairness in comparison, the model is trained on the complete training set of seen classes instead of 16 shots per class. 
In this case, we extract the image feature from CLIP visual encoder and obtain the corresponding class attribute from the prompt template ``a photo of a \{class\}".
As in~\cite{xian2018zero}, we report the average per-class top-1 accuracy on seen  and unseen  classes. 
Furthermore, the harmonic mean is also reported to provide a balance between seen and unseen accuracy.

\noindent\textbf{Results.}
The results of generalized zero-shot learning are shown in Table~\ref{tab:gzsl}.
Experiments are conducted on three standard benchmarks for zero-shot classification: CUB~\cite{welinder2010caltech}, AWA2~\cite{xian2018zero}, and FLO~\cite{nilsback2008automated}.
We choose f-CLSWGAN~\cite{xian2018feature}, Cycle-WGAN~\cite{felix2018multi}, LisGAN~\cite{li2019leveraging}, TCN~\cite{jiang2019transferable}, f-VAEGAN~\cite{xian2019f}, TF-VAEGAN~\cite{narayan2020latent}, GCM-CF~\cite{yue2021counterfactual}, HSVA~\cite{chen2021hsva}, and MSDN~\cite{chen2022msdn} as our baseline methods.
These methods extract the average-pooled feature instances of size 2,048 from the ImageNet-1K~\cite{deng2009imagenet} pretrained ResNet-101~\cite{he2016deep}.
And they use expert annotated attributes or text descriptions~\cite{mikolov2013distributed} as auxiliary information of classes, which requires additional human labor.

The results reported in Table~\ref{tab:gzsl} indicate that CoOp~\cite{zhou2022learning} yields a substantial improvement in the performance of seen classes. 
Specifically, the method leads to a 9.0\%, 2.2\%, and 17.9\% performance increase on CUB, AWA2, and FLO datasets, respectively.
However, the performance of CoOp on unseen classes is comparatively lower, as evidenced by a decline of 6.0\%, 15.6\%, and 13.4\% on CUB, AWA2, and FLO datasets, respectively, compared to CLIP~\cite{radford2021learning}.
This observation suggests that CoOp may suffer from severe overfitting on the seen classes.
In this regard, our proposed method, CoOp + SHIP, leverages generative prompt tuning to enhance the performance of unseen classes.
Our experimental results demonstrate that CoOp + SHIP leads to significant gains of +6.1\%, +11.4\%, and +16.8\% on unseen classes compared to CoOp~\cite{zhou2022learning}. 
Furthermore, the performance of CoOp + SHIP is comparable or superior to previous zero-shot learning methods.

To ensure a fair comparison, we have implemented the TF-VAEGAN~\cite{narayan2020latent} and f-VAEGAN~\cite{xian2019f} using CLIP extracted features, with the attribute of each class generated through the prompt template ``a photo of a \{class\}".
The results presented in the table indicate that while these models achieve the highest performance on seen classes, their performance on unseen classes is significantly lower, suggesting that these models suffer from severe overfitting to the seen classes.
We presume that the use of a coarse prompt template such as ``a photo of a \{class\}" may not provide sufficient transferability compared to expert-annotated attributes used in previous methods.

\subsection{Ablation Study}

\noindent\textbf{Different generative models.}
We conducted a series of experiments to investigate the effectiveness of the generative framework and the CLIP-based generator.
For this, we implemented four distinct types of generators, with two types of frameworks and two types of generators.
Table~\ref{tab:generator} presents the experimental results. 
In the table, G denotes the use of GAN~\cite{arjovsky2017wasserstein} as the framework, while V denotes the use of VAE~\cite{kingma2013auto} as the framework.
S represents the training of a three-layer MLP as a generator from scratch, while T denotes the utilization of the CLIP-based generator discussed in Section~\ref{sec:method}.
Notably, V + T is equivalent to our model.

\begin{table}[htbp]
  \centering
  \caption{We conducted an ablation analysis to evaluate the effectiveness of the generative frameworks and generators. 
  We add them to CoOp and the results are average on the 11 datasets.
  The last row is CoOp's results. G: use a GAN-based framework. V: use a VAE-based framework. S: train the generator from scratch. T: using the CLIP-based generator.}
    \begin{tabular}{cc|ccc}
    \toprule
    \textbf{framework} & \textbf{generator} & base  & new   & H \\
    \midrule
    G     & S     & 79.96  & 59.35  & 67.30  \\
    V     & S     & 79.05  & 69.32  & 73.41  \\
    G     & T     & 77.69  & 67.32  & 71.68  \\
    V     & T     & 80.03  & \textbf{73.69}  & \textbf{76.73}  \\
    -     & -     & \textbf{82.69}  & 63.22  & 71.66  \\
    \bottomrule
    \end{tabular}%
  \label{tab:generator}%
\end{table}%

We incorporate the generative models\ into CoOp~\cite{zhou2022learning} and evaluate their performance on the 11 datasets mentioned above.
In Table~\ref{tab:generator}, the results indicate that VAE-based models outperform GAN-based models, supporting our claims that GANs~\cite{goodfellow2020generative} are difficult to train with the few-shot base dataset, leading to a suboptimal performance on new classes.
Additionally, we find that utilizing the CLIP-based generator yields superior results to straightforwardly training the generator from scratch, highlighting the effectiveness of our CLIP-based generator and the efficient utilization of pretrained knowledge of CLIP.
Furthermore, the table reveals that the combination of CoOp with G + S yields inferior performance compared to vanilla CoOp~\cite{zhou2022learning}.
This indicates that not arbitrary data generation for new classes can improve model performance.
Based on these results, we select VAE~\cite{kingma2013auto} as our generative architecture and choose to utilize the CLIP-based generator.

\begin{table}[htbp]
  \centering
  \caption{
  We evaluate different forms of prompts. 
  Results are average on the 11 datasets. 
  \textbf{global} denotes whether using global prompts. 
  And \textbf{sequential} denotes whether the local bias is sequential or identical.}
    \begin{tabular}{cc|ccc}
    \toprule
    \textbf{global} & \textbf{sequential} & base  & new   & H \\
    \midrule
    \ding{55}     & \ding{55}     & 80.70 & 71.60 & 75.37 \\
    \ding{51}     & \ding{55}     & 80.03 & \textbf{73.69} & \textbf{76.73} \\
    \ding{55}     & \ding{51}     & \textbf{80.77} & 71.95 & 75.73 \\
    \ding{51}     & \ding{51}     & 80.59 & 70.75 & 74.89  \\
    \bottomrule
    \end{tabular}%
  \label{tab:prompt}%
\end{table}%

\begin{figure*}[htbp]     
    \centering
    (a) 
    \hspace{0.3cm}
    \begin{minipage}{4.5cm}
        \centering``aerial - aerial - \\
        \centering aerial - aerial.'' \\
        \centering (EuroSAT)
    \end{minipage}
    \hspace{0.3cm}
    \begin{minipage}[c]{0.15\textwidth}
    \includegraphics[height=0.8in]{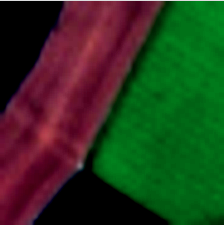}
    \end{minipage}
    (b) 
    \hspace{0.3cm}
    \begin{minipage}{4.5cm}
        ``swift-footed - swift-footed - \\
        swift-footed - swift-footed.'' \\
        \centering (OxfordPets)
    \end{minipage}
    \hspace{0.3cm}
    \begin{minipage}[c]{0.15\textwidth}
    \includegraphics[height=0.8in]{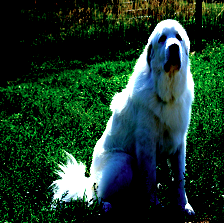}
    \end{minipage}
    \\
    (c) 
    \hspace{0.3cm}
    \begin{minipage}{4.5cm}
        \centering ``fierce - fierce - \\
        \centering polydactyl - polydactyl.'' \\
        \centering (OxfordPets)
    \end{minipage}
    \hspace{0.3cm}
    \begin{minipage}[c]{0.15\textwidth}
    \includegraphics[height=0.8in]{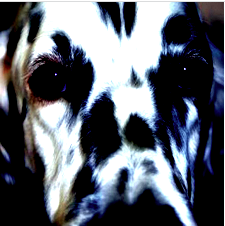}
    \end{minipage}
    (d) 
    \hspace{0.3cm}
    \begin{minipage}{4.5cm}
        ``\centering outstretched - odd-pinnate - \\
        \centering odd-pinnate - three-lobed.'' \\
        \centering (Flowers102)
    \end{minipage}
    \hspace{0.3cm}
    \begin{minipage}[c]{0.15\textwidth}
    \includegraphics[height=0.8in]{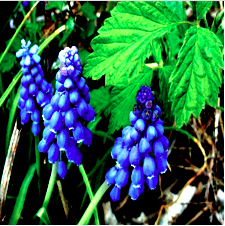}
    \end{minipage}
    \\
    (e) 
    \hspace{0.3cm}
    \begin{minipage}{4.5cm}
        ``\centering outstretched - outstretched - \\
        \centering outstretched - short-stem.'' \\
        \centering (Flowers102)
    \end{minipage}
    \hspace{0.3cm}
    \begin{minipage}[c]{0.15\textwidth}
    \includegraphics[height=0.8in]{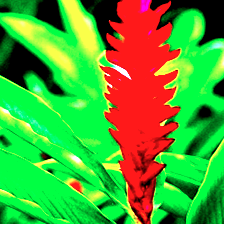}
    \end{minipage}
    (f) 
    \hspace{0.3cm}
    \begin{minipage}{4.5cm}
        ``\centering vibrational - odd-pinnate - \\
        \centering odd-pinnate - odd-pinnate.'' \\
        \centering (Flowers102)
    \end{minipage}
    \hspace{0.3cm}
    \begin{minipage}[c]{0.15\textwidth}
    \includegraphics[height=0.8in]{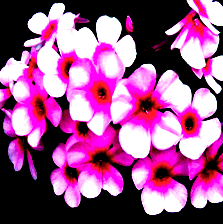}
    \end{minipage}
    \caption{Interpretation of prompts in the latent space.
    We observe that some words are capable of characterizing the attributes present in the images.
    However, since we utilize the identical bias in the prompts, some words are the same.}
    \label{fig:interpretation}
\end{figure*}

\noindent\textbf{Different forms of generative prompts.}
The prompts used in our method comprise a fixed set of global prompts and a local instance-specific bias, as described in Section~\ref{sec:method}.
More specifically, the prompts are represented as an addition of the global prompts and the local bias, i.e., $\bm{p} = [\bm{p_1} + \bm{r}, \bm{p_2} + \bm{r}, ... \bm{p_L} + \bm{r}]$. 
We investigate the impact of different forms of prompts on performance.
We use the term \textbf{global} to denote the use of global prompts, $[\bm{p_1}, \bm{p_2}, ... \bm{p_L}]$, and the term \textbf{sequential} to refer to the use of identical or sequential local bias, i.e., $[\bm{r}, \bm{r}, ..., \bm{r}]$ or $[\bm{r}_1, \bm{r}_2, ..., \bm{r}_L]$.
The results, presented in Table~\ref{tab:prompt}, indicate that utilizing global prompts along with identical local bias yields the best performance.
And using local prompts alone (whether sequential or not) results in a negative impact on the new class performance, underscoring the importance of global prompts in capturing vital information.
Notably, using both \textbf{global} and \textbf{sequential} prompts results in the worst performance.
This may be attributed to the instability of training since they both learn sequential prompts.

\noindent\textbf{Interpretation of prompts.}
One benefit of our CLIP-based generative model is that we can provide interpretive prompts.
The model learns the mapping from visual features to token embedding space via the VAE process.
By utilizing this mapping, we can obtain instance-specific prompts for the input image.
The next step involves selecting the nearest natural words from the vocabulary based on their Euclidean distance to the prompts in latent space.
However, the approach maps continuous vectors into discrete codes of words, which can result in generated sentences that may not necessarily be semantically coherent, as noted in prior research~\cite{zhou2022learning}.

The interpretation of the prompts reveals several noteworthy observations.
As depicted in Figure~\ref{fig:interpretation} (a), the model has learned to associate the term `aerial" with images captured from an aerial perspective in EuroSAT.
Furthermore, the model has accurately identified some characteristics of dogs, as exemplified in Figure~\ref{fig:interpretation} (b)-(c), where the terms ``swift-footed" and ``fierce" can be used to describe animals.
Additionally, the model has demonstrated an understanding of floral morphology, as demonstrated in Figure~\ref{fig:interpretation} (d)-(f), where the terms ``odd-pinnate," ``three-lobed," and ``shot-stem" are employed to describe characteristics of flowers. 
Since we utilize the identical local bias in the prompts, some words are the same in the interpretation sentence.

Although the interpretation may not be entirely precise, it provides valuable insights into the images.
We hope the results inform future studies on interpretable vision-language inference and yield further insights.
 
\section{Conclusion}
In this paper, we provide a generative approach, SHIP, to handle the scenario where some classes have no data. 
By training a data-efficient generator to bridge the data gap in new classes, we improve CLIP performance on various tasks using off-the-shelf methods, including base-to-new generalization, cross-data transfer learning, and generalized zero-shot classification.
Although achieving remarkable results, it requires additional training costs, which we aim to mitigate in future research. Additionally, future work will explore the applicability of SHIP in dense prediction tasks.


{\small
\bibliographystyle{ieee_fullname}
\bibliography{egbib}
}

\newpage
\appendix
\renewcommand{\appendixname}{Appendix~\Alph{section}}
\section{Appendix}

\subsection{Datasets details}
The details of the 11 datasets used in base-to-new generalization and cross-dataset transfer learning are shown in Table~\ref{tab:b2nd}. 
In addition, the statistic of datasets used in generalized zero-shot learning is summarized in Table~\ref{tab:gzsld}.

\begin{table}[htbp]
  \centering
  \caption{Datasets statistic of 11 datasets for base-to-new generalization and cross-dataset transfer learning.}
  \resizebox{0.48\textwidth}{!}{
    \begin{tabular}{llrrrr}
    \toprule
    \multicolumn{2}{l}{Dataset} & classes & train & val   & test \\
    \midrule
    \multicolumn{2}{l}{ImageNet~\cite{deng2009imagenet}} & 1,000 & 1.28M & N/A   & 50,000 \\
    \multicolumn{2}{l}{Caltech101~\cite{fei2004learning}} & 100   & 4,128 & 1,649 & 2,465 \\
    \multicolumn{2}{l}{OxfordPets~\cite{parkhi2012cats}} & 37    & 2,944 & 736   & 3,669 \\
    \multicolumn{2}{l}{StanfordCars~\cite{krause20133d}} & 196   & 6,509 & 1,635 & 8,041 \\
    \multicolumn{2}{l}{Flowers102~\cite{nilsback2008automated}} & 102   & 4,093 & 1,633 & 2,463 \\
    \multicolumn{2}{l}{Food101~\cite{bossard2014food}} & 101   & 50,500 & 20,200 & 30,300 \\
    \multicolumn{2}{l}{FGVCAircraft~\cite{maji2013fine}} & 100   & 3,334 & 3,333 & 3,333 \\
    \multicolumn{2}{l}{SUN397~\cite{xiao2010sun}} & 397   & 15,880 & 3,970 & 19,850 \\
    \multicolumn{2}{l}{DTD~\cite{cimpoi2014describing}} & 47    & 2,820 & 1,128 & 1,692 \\
    \multicolumn{2}{l}{EuroSAT~\cite{helber2019eurosat}} & 10    & 13,500 & 5,400 & 8,100 \\
    \multicolumn{2}{l}{UCF101~\cite{soomro2012ucf101}} & 101   & 7,639 & 1,898 & 3,783 \\
    \bottomrule
    \end{tabular}%
   }
  \label{tab:b2nd}%
\end{table}%

\begin{table}[htbp]
  \centering
  \caption{Datasets statistic of datasets in generalized zero-shot learning.}
  \resizebox{0.48\textwidth}{!}{
    \begin{tabular}{lllrrr}
    \toprule
    \multicolumn{3}{l}{Dataset} & CUB~\cite{welinder2010caltech}  & AWA2~\cite{xian2018zero}  & FLO~\cite{nilsback2008automated} \\
    \midrule
    \multicolumn{3}{l}{\# of Attributes} & 312   & 85    & 1,024 \\
    \multicolumn{3}{l}{\# of seen classes} & 150   & 40    & 82 \\
    \multicolumn{3}{l}{\# of unseen classes} & 50    & 10    & 20 \\
    \multicolumn{3}{l}{\# of total images} & 11,788 & 30,475 & 8,189 \\
    \bottomrule
    \end{tabular}%
   }
  \label{tab:gzsld}%
\end{table}%

\subsection{Generalized zero-shot setting}
The current evaluation protocol utilized in base-to-new generalization assumes that base and new classes are completely isolated during testing, which may not reflect a realistic scenario. In contrast, in a more realistic setting, test sets contain a mix of base and new class data, as previously employed in generalized zero-shot learning. We refer to this as the generalized zero-shot setting and re-evaluate base-to-new generalization under this setting.
The results of our evaluation are presented in Table~\ref{tab:b2ngzsl}, which indicates a significant decrease in performance for previous methods such as CoOp~\cite{zhou2022learning} and CLIP-Adapter~\cite{gao2021clip} under this more strict setting.
Conversely, our proposed method, SHIP, continues to improve performance in new classes.

\subsection{Different lengths of prompts.}
As described in Sec.~\ref{sec:method}, our proposed approach generates instance-specific prompts to produce corresponding features, which consist of a global prompt and a local bias. Specifically, the prompts is computed as follows:
\begin{equation}
    \bm{p} = [\bm{p}_1 + \bm{r}, \bm{p}_2 + \bm{r}, ..., \bm{p}_L + \bm{r}],
\end{equation}
where $L$ is the length of prompts.
To examine the influence of prompt length on our method's performance, we conduct an ablation study, the results of which are presented in Table~\ref{tab:diffl}.
Specifically, we set the prompt lengths in our approach to 1, 2, 4, and 8 and integrate our method into CoOp~\cite{zhou2022learning} to evaluate its performance on base-to-new generalization. 
Our experimental results indicate that our proposed approach performs best when the prompt length is set to $L=4$. Therefore, we set the default prompt length as $L=4$ for our experiments.

\begin{table}[htbp]
  \centering
  \caption{
  We evaluate different lengths of prompts. 
  Results are average on the 11 datasets. 
  }
    \begin{tabular}{ccccc}
    \toprule
    \multicolumn{2}{c}{} & Base  & New   & H \\
    \midrule
    \multicolumn{2}{c}{CoOp + SHIP (L=1)} & 80.61  & 71.69  & 75.53  \\
    \multicolumn{2}{c}{CoOp + SHIP (L=2)} & \textbf{80.61} & 70.28  & 74.56  \\
    \multicolumn{2}{c}{CoOp + SHIP (L=4)} & 80.03  & \textbf{73.69} & \textbf{76.73} \\
    \multicolumn{2}{c}{CoOp + SHIP (L=8)} & 80.54  & 72.05  & 75.70  \\
    \bottomrule
    \end{tabular}%
  \label{tab:diffl}%
\end{table}%

\setlength{\tabcolsep}{10.0pt}
\begin{table*}[htbp]
  \centering
  \caption{Evaluate base-to-new generalization under the generalized zero-shot setting, where the base and new data are mixed together in the test dataset.}
  \resizebox{1.0\textwidth}{!}{
    \begin{tabular}{ll|rrr|rrr|rrr|rrr}
    \toprule
    \multicolumn{2}{c|}{\multirow{2}[2]{*}{}} & \multicolumn{3}{c|}{\textbf{Average}} & \multicolumn{3}{c|}{\textbf{ImageNet~\cite{deng2009imagenet}}} & \multicolumn{3}{c|}{\textbf{Caltech101~\cite{fei2004learning}}} & \multicolumn{3}{c}{\textbf{OxfordPets~\cite{parkhi2012cats}}} \\
    \multicolumn{2}{c|}{} & \multicolumn{1}{c}{\textbf{Base}} & \multicolumn{1}{c}{\textbf{New}} & \multicolumn{1}{c|}{\textbf{H}} & \multicolumn{1}{c}{\textbf{Base}} & \multicolumn{1}{c}{\textbf{New}} & \multicolumn{1}{c|}{\textbf{H}} & \multicolumn{1}{c}{\textbf{Base}} & \multicolumn{1}{c}{\textbf{New}} & \multicolumn{1}{c|}{\textbf{H}} & \multicolumn{1}{c}{\textbf{Base}} & \multicolumn{1}{c}{\textbf{New}} & \multicolumn{1}{c}{\textbf{H}} \\
    \midrule
    \multicolumn{2}{l|}{CLIP} & 63.37  & \textbf{67.39} & 65.32  & 70.04  & 67.52  & 68.76  & 93.61  & \textbf{91.70} & 92.65  & 84.90  & \textbf{93.51} & 89.00  \\
    \multicolumn{2}{l|}{CoOp} & 79.76  & 46.91  & 59.08  & 72.51  & \textbf{67.70} & \textbf{70.02} & 97.68  & 85.04  & 90.92  & 93.67  & 64.99  & 76.74  \\
    \multicolumn{2}{l|}{CoOp + SHIP} & 78.74  & 58.53  & \textbf{67.15} & 71.96  & 67.12  & 69.45  & 96.45  & 90.07  & \textbf{93.15} & \textbf{94.58} & 89.15  & \textbf{91.78} \\
    \multicolumn{2}{l|}{CLIP-Adapter} & \textbf{82.79} & 30.48  & 44.55  & 71.94  & 64.95  & 68.27  & 98.06  & 71.07  & 82.41  & 91.65  & 33.33  & 48.89  \\
    \multicolumn{2}{l|}{CLIP-Adapter  + SHIP} & 82.53  & 35.73  & 49.87  & 72.02  & 66.17  & 68.97  & 97.61  & 77.18  & 86.20  & 91.81  & 40.83  & 56.52  \\
    \multicolumn{2}{l|}{Tip-Adapter + SHIP} & 82.07  & 49.02  & 61.38  & \textbf{75.46} & 60.80  & 67.34  & \textbf{98.26} & 81.55  & 89.13  & 93.99  & 83.17  & 88.25  \\
    \bottomrule    
    \end{tabular}%
  }
    \vspace{0.1cm}
    \resizebox{1.0\textwidth}{!}{
    \begin{tabular}{ll|rrr|rrr|rrr|rrr}
    \toprule
    \multicolumn{2}{c|}{\multirow{2}[2]{*}{}} & \multicolumn{3}{c|}{\textbf{StanfordCars~\cite{krause20133d}}} & \multicolumn{3}{c|}{\textbf{Flowers102~\cite{nilsback2008automated}}} & \multicolumn{3}{c|}{\textbf{Food101~\cite{bossard2014food}}} & \multicolumn{3}{c}{\textbf{FGVCAircraft~\cite{maji2013fine}}} \\
    \multicolumn{2}{c|}{} & \multicolumn{1}{c}{\textbf{Base}} & \multicolumn{1}{c}{\textbf{New}} & \multicolumn{1}{c|}{\textbf{H}} & \multicolumn{1}{c}{\textbf{Base}} & \multicolumn{1}{c}{\textbf{New}} & \multicolumn{1}{c|}{\textbf{H}} & \multicolumn{1}{c}{\textbf{Base}} & \multicolumn{1}{c}{\textbf{New}} & \multicolumn{1}{c|}{\textbf{H}} & \multicolumn{1}{c}{\textbf{Base}} & \multicolumn{1}{c}{\textbf{New}} & \multicolumn{1}{c}{\textbf{H}} \\
    \midrule
    \multicolumn{2}{l|}{CLIP~\cite{radford2021learning}} & 59.75  & \textbf{70.96} & 64.87  & 68.00  & \textbf{73.90} & 70.83  & 85.84  & \textbf{86.39} & \textbf{86.11} & 19.33  & \textbf{29.87} & \textbf{23.47} \\
    \multicolumn{2}{l|}{CoOp~\cite{zhou2022learning}} & 69.24  & 60.01  & 64.30  & 94.40  & 37.45  & 53.62  & 88.73  & 78.05  & 83.05  & 36.07  & 13.80  & 19.96  \\
    \multicolumn{2}{l|}{CoOp + SHIP} & 67.39  & 67.07  & \textbf{67.23} & 94.97  & 61.42  & \textbf{74.59} & 87.82  & 83.83  & 85.78  & 33.43  & 16.80  & 22.36  \\
    \multicolumn{2}{l|}{CLIP-Adapter~\cite{gao2021clip}} & 79.26  & 34.36  & 47.94  & \textbf{98.38} & 27.66  & 43.18  & 88.42  & 44.51  & 59.21  & \textbf{42.50} & 8.58  & 14.28  \\
    \multicolumn{2}{l|}{CLIP-Adapter + SHIP} & 78.46  & 39.07  & 52.16  & 97.72  & 34.26  & 50.73  & 88.31  & 51.70  & 65.22  & 42.20  & 10.26  & 16.50  \\
    \multicolumn{2}{l|}{Tip-Adapter + SHIP} & \textbf{79.81} & 51.84  & 62.86  & 95.35  & 35.25  & 51.47  & \textbf{89.84} & 75.69  & 82.16  & 41.54  & 14.58  & 21.58  \\
    \bottomrule
    \end{tabular}%
  }
    \vspace{0.1cm}
    \resizebox{1.0\textwidth}{!}{
    \begin{tabular}{ll|rrr|rrr|rrr|rrr}
    \toprule
    \multicolumn{2}{c|}{\multirow{2}[2]{*}{}} & \multicolumn{3}{c|}{\textbf{SUN397~\cite{xiao2010sun}}} & \multicolumn{3}{c|}{\textbf{DTD~\cite{cimpoi2014describing}}} & \multicolumn{3}{c|}{\textbf{EuroSAT~\cite{helber2019eurosat}}} & \multicolumn{3}{c}{\textbf{UCF101~\cite{soomro2012ucf101}}} \\
    \multicolumn{2}{c|}{} & \multicolumn{1}{c}{\textbf{Base}} & \multicolumn{1}{c}{\textbf{New}} & \multicolumn{1}{c|}{\textbf{H}} & \multicolumn{1}{c}{\textbf{Base}} & \multicolumn{1}{c}{\textbf{New}} & \multicolumn{1}{c|}{\textbf{H}} & \multicolumn{1}{c}{\textbf{Base}} & \multicolumn{1}{c}{\textbf{New}} & \multicolumn{1}{c|}{\textbf{H}} & \multicolumn{1}{c}{\textbf{Base}} & \multicolumn{1}{c}{\textbf{New}} & \multicolumn{1}{c}{\textbf{H}} \\
    \midrule
    \multicolumn{2}{l|}{CLIP~\cite{radford2021learning}} & 60.34  & \textbf{64.91} & 62.54  & 42.13  & \textbf{46.86} & \textbf{44.37} & 49.81  & \textbf{45.44} & \textbf{47.52} & 63.34  & \textbf{70.20} & 66.59  \\
    \multicolumn{2}{l|}{CoOp~\cite{zhou2022learning}} & 78.75  & 43.44  & 56.00  & 72.80  & 14.86  & 24.68  & 90.81  & 0.64  & 1.27  & 82.68  & 50.08  & 62.38  \\
    \multicolumn{2}{l|}{CoOp + SHIP} & 75.49  & 59.21  & \textbf{66.37} & 75.23  & 27.78  & 40.57  & 90.67  & 16.74  & 28.27  & 78.18  & 64.63  & \textbf{70.76} \\
    \multicolumn{2}{l|}{CLIP-Adapter~\cite{gao2021clip}} & \textbf{79.06} & 20.35  & 32.37  & 81.94  & 2.90  & 5.60  & \textbf{93.60} & 0.05  & 0.10  & \textbf{85.83} & 27.47  & 41.62  \\
    \multicolumn{2}{l|}{CLIP-Adapter + SHIP} & 78.96  & 27.63  & 40.93  & \textbf{82.29} & 7.85  & 14.33  & 93.05  & 0.44  & 0.87  & 85.42  & 37.59  & 52.20  \\
    \multicolumn{2}{l|}{Tip-Adapter + SHIP} & 76.60  & 51.80  & 61.80  & 79.98  & 15.82  & 26.42  & 89.83  & 10.41  & 18.66  & 82.11  & 58.30  & 68.19  \\
    \bottomrule
    \end{tabular}%
  }
  \label{tab:b2ngzsl}%
\end{table*}%
\end{document}